\documentclass{llncs} %

\usepackage{float}
\usepackage{graphicx}
\usepackage{wrapfig}
\usepackage{times}

\usepackage[labelfont=bf]{caption}
\usepackage{amsmath} 
\usepackage{mathtools} 
\usepackage{amsfonts} 
\usepackage{amssymb}
\usepackage{comment}

\usepackage[table,xcdraw,svgnames]{xcolor}

\usepackage{array}
\usepackage{booktabs}
\usepackage{rotating}
\usepackage{multirow}
\usepackage{multicol}
\usepackage{lscape}
\usepackage{subfig}

\definecolor{babyblue}{rgb}{0.54, 0.81, 0.94}
\definecolor{armygreen}{rgb}{0.29, 0.33, 0.13}
\definecolor{brightlavender}{rgb}{0.75, 0.58, 0.89}
\definecolor{aqua}{rgb}{0.0, 1.0, 1.0}
\definecolor{caribbeangreen}{rgb}{0.0, 0.8, 0.6}
\definecolor{reddish}{rgb}{0.82, 0.1, 0.26}
\definecolor{emerald}{rgb}{0.31, 0.78, 0.47}
\definecolor{jasper}{rgb}{0.84, 0.23, 0.24}
\definecolor{red}{rgb}{1.0, 0.0, 0.0}
\definecolor{green}{rgb}{0.0, 1.0, 0.0}
\definecolor{blue}{rgb}{0.0, 0.0, 1.0}
\definecolor{darkgreen}{rgb}{0.1, 0.7, 0.1}
\definecolor{darkblue}{rgb}{0.1, 0.1, 0.7}
\definecolor{red}{rgb}{0.7, 0.1, 0.1}

\usepackage{tikz}
\usetikzlibrary{shadows}
\usepackage[export]{adjustbox}
\usepackage[colorlinks=true,linkcolor=caribbeangreen,citecolor=emerald]{hyperref}
\usepackage[colorinlistoftodos,prependcaption,textsize=tiny]{todonotes}
\usepackage[align=center,shadow=true,shadowsize=5pt,nobreak=true,framemethod=tikz,skipabove=2pt,skipbelow=1pt,innertopmargin=-3pt,innerbottommargin=3pt,innerleftmargin=5pt,innerrightmargin=5pt,leftmargin=-2pt,rightmargin=-2pt]{mdframed}

\usepackage{algorithm,algorithmicx,algpseudocode}

\usepackage{color,soul}

\usepackage{tikz,xcolor,hyperref}

\definecolor{lime}{HTML}{A6CE39}
\DeclareRobustCommand{\orcidicon}{
	\begin{tikzpicture}
	\draw[lime, fill=lime] (0,0) 
	circle [radius=0.16] 
	node[white] {{\fontfamily{qag}\selectfont \tiny ID}};
	\draw[white, fill=white] (-0.0625,0.095) 
	circle [radius=0.007];
	\end{tikzpicture}
	\hspace{-2mm}
}

\foreach \x in {A, ..., Z}{\expandafter\xdef\csname orcid\x\endcsname{\noexpand\href{https://orcid.org/\csname orcidauthor\x\endcsname}
			{\noexpand\orcidicon}}
}
			
\begin{document}

\title{DuoGNN: Topology-aware Graph Neural Network with Homophily and Heterophily Interaction-Decoupling}

\titlerunning{Short Title}  

\author{Kevin Mancini \index{Mancini, Kevin}  \and Islem Rekik\orcidA{}\index{Rekik, Islem}\thanks{ {corresponding author: i.rekik@imperial.ac.uk, \url{http://basira-lab.com}, GitHub: \url{http://github.com/basiralab}. }}}

\institute{BASIRA Lab, Imperial-X (I-X) and Department of Computing, Imperial College London, London, United Kingdom}

\authorrunning{}

\maketitle              

\begin{abstract}

Graph Neural Networks (GNNs) have proven effective in various medical imaging applications, such as automated disease diagnosis. However, due to the local neighborhood aggregation paradigm in message passing which characterizes these models, they inherently suffer from two fundamental limitations: \emph{first}, indistinguishable node embeddings due to heterophilic node aggregation (known as over-smoothing), and \emph{second}, impaired message passing due to aggregation through graph bottlenecks (known as over-squashing). These challenges hinder the model expressiveness and prevent us from using deeper models to capture long-range node dependencies within the graph. Popular solutions in the literature are either too expensive to process large graphs due to high time complexity or do not generalize across all graph topologies. To address these limitations, we propose DuoGNN, a scalable and generalizable architecture which leverages topology to decouple homophilic and heterophilic edges and capture both short-range and long-range interactions. Our three core contributions introduce (i) a topological edge-filtering algorithm which extracts homophilic interac- tions and enables the model to generalize well for any graph topology, (ii) a heterophilic graph condensation technique which extracts heterophilic interactions and ensures scalability, and (iii) a dual homophilic and heterophilic aggregation pipeline which prevents over-smoothing and over-squashing during the message passing. We benchmark our model on medical and non-medical node classification datasets and compare it with its variants, showing consistent improvements across all tasks. Our DuoGNN code is available at \href{https://github.com/basiralab/DuoGNN}{https://github.com/basiralab/DuoGNN}.

\end{abstract}


\section{Introduction}

GNNs \cite{scarselli2008graph} are a machine learning model designed to process graph-structured data (e.g., molecule structure \cite{NPMS}, and protein interactions \cite{PPI}). GNNs have proven effective in various graph-based downstream \cite{xu2018powerful} tasks such as node classification, graph classification, link prediction \cite{surveyGNN}, and time-series prediction\cite{cui2021metro}. Recently, they also have demonstrated efficacy in medical fields \cite{GCNmedical} such as neuroscience \cite{basira-neuroscience}, and image segmentation for medical purposes (e.g., liver tumour and colon pathology classification  \cite{medmnistv2}).

\begin{wrapfigure}{r}{0.5\textwidth}
  \vspace{-25pt} 
  \centering
  \includegraphics[width=\linewidth]{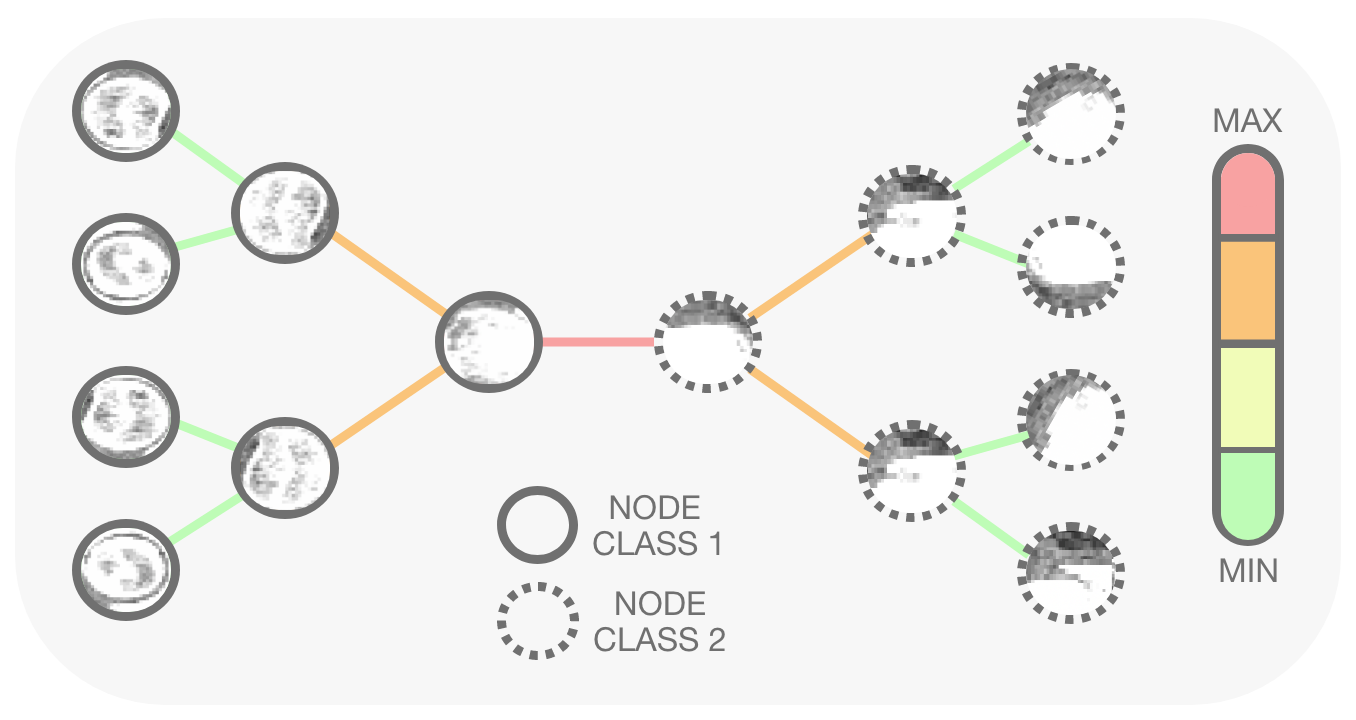}
  \caption{Bottleneck in a graph. The nodes are samples from OrganSMNIST of two distinct classes and the edges connect similar nodes. The edge color indicates the variation of the receptor field, which increases as nodes get closer to the bottleneck.}
  \label{fig:bottleneck}
\end{wrapfigure}

The strength of GNN models lies in their ability to process graph-structured data and capture short-range spatial interactions between the nodes through local neighbourhood aggregation in the message passing. However, this local aggregation paradigm may be ineffective on specific graph densities and structures. For instance, when processing a strongly clustered graph, a standard GNN model will likely fail to capture interactions between nodes belonging to distant clusters. This is an extensively studied limitation of GNNs and theoretical findings have formalized the root causes of this phenomenon as over-smoothing and over-squashing \cite{qureshi2023limits}. \emph{Over-smoothing} (see \textbf{Def}~\ref{over-smoothing}) is often described as the phenomenon where the features of nodes belongings to distinct classes become undistinguishable as the number of layers in the GNN increases \cite{rusch2023survey}.  \emph{Over-squashing} (see \textbf{Def}~\ref{over-squashing}), on the other hand, is the inhibition of the message-passing capabilities of the graph caused by graph bottlenecks (\textbf{Fig}~\ref{fig:bottleneck}). 

These phenomena limit the performance in many applications and prevent us from using deep GNNs to capture long-range dependencies. To address this challenge, recent works proposed enhanced GNN models by integrating graph transformer-based modules (i.e, global attention \cite{fei2023gnn}, local and global attention \cite{wu2021representing}), pre-processing techniques (i.e, curvature-enhanced edge rewiring \cite{nguyen2023revisiting}, topological rewiring \cite{barbero2023locality}) and multi-scale architectures \cite{dong2023megraph,chen2023multi}. Although these methods can marginally enhance the model performance, they fail to propose a generalized and scalable approach which can process large graphs and be effective in any graph topology. Specifically, attention-based approaches are constrained by their quadratic time complexity and lack of topological awareness. Rewiring algorithms are not powerful enough for very large graphs and still require deep networks that inevitably lead to over-smoothing. Multi-scale solutions introduce considerable additional time complexity and do not generalize well to various graph topologies, as they primarily process clustered graphs (i.e., Stochastic Block Models). For these reasons, a scalable and general method for learning long-range interactions (LRIs) would significantly expand the applicability of GNNs in medical imaging applications. Such a method would be a key contribution to the field, as medical imaging tasks often involve large datasets and require capturing complex patterns that may span distant nodes.

\begin{figure}[tb]
\centering
\includegraphics[width = 0.73\hsize]{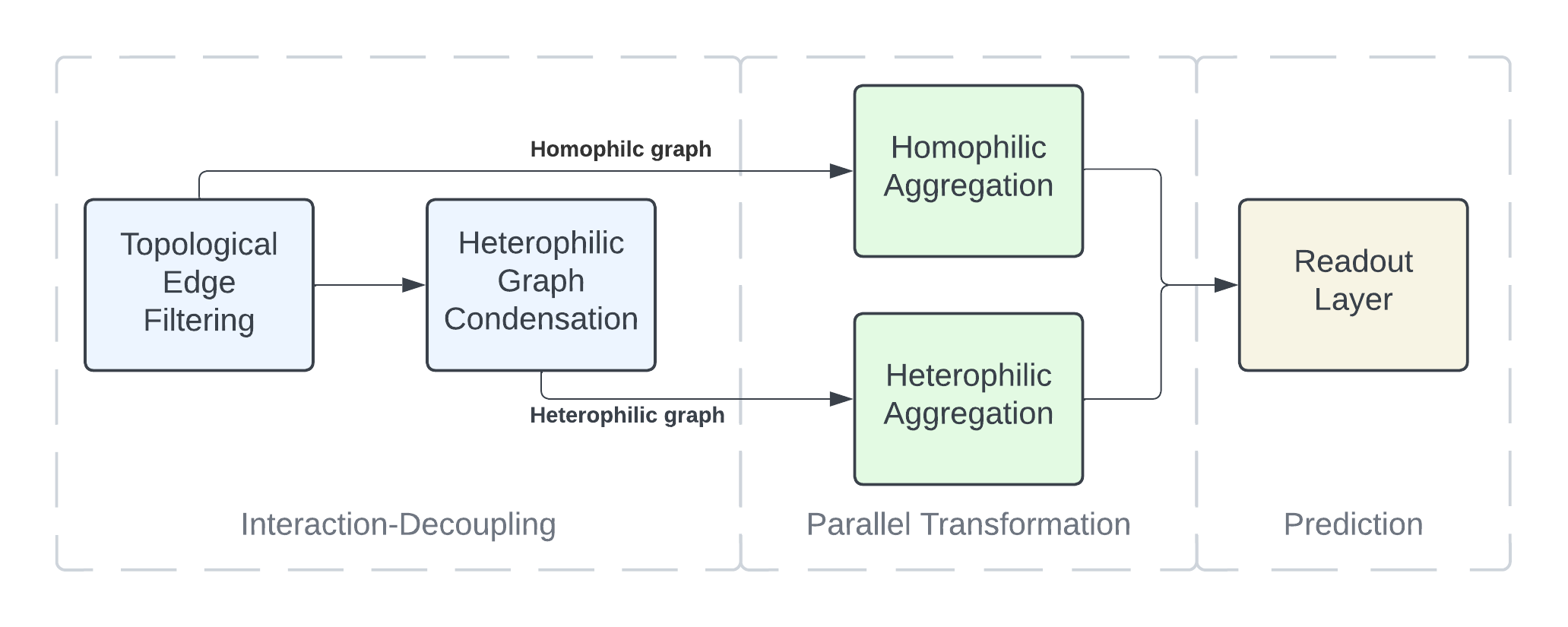}
\caption{DuoGNN's three-stage architecture pipeline}
\label{fig:flow_model}
\end{figure}

To the best of our knowledge, current GNN models lack a generalizable and scalable solution for \emph{simultaneously} capturing short-range and long-range node dependencies. To address these limitations, we propose DuoGNN, a topology-aware architecture that leverages topological graph properties to efficiently capture short-range and long-range interactions in any graph structure and density. The core idea behind our model is to distinguish between \emph{homophilic} and \emph{heterophilic} (\textbf{Def}~\ref{homphily}) node interactions during a topological \emph{interaction-decoupling} stage and to process them independently through a \emph{parallel transformation} stage (\textbf{Fig}~\ref{fig:flow_model}). This process prevents the model from suffering from over-smoothing and over-squashing. Finally, we benchmark our model using both medical and non-medical datasets, comparing the results with SOTA methods.

\section{Related work}
\textbf{Transformer-based self-attention.}
A common approach to overcome over-smoothing and over-squashing and capture long-range node interactions is integrating a transformer-based module to the GNN (i.e, global attention \cite{fei2023gnn}, local and global attention \cite{wu2021representing}). Transformers can aggregate information globally without being limited by the local neighborhood aggregation paradigm, making them a very effective solution. However, they fail to propose a scalable solution which can process large-scale graphs, which is arguably the scenario where long-range interaction detection is of the utmost importance. Graph self-attention has a time complexity of \( O(|\mathcal{V}|^2) \), where \(|\mathcal{V}|\) refers to the number of nodes in the graph, this complexity is unsuitable for large graphs. Moreover, classical transformers are inefficient at processing LRIs in graphs because they are inherently topology-agnostic and consequently process all \( |\mathcal{V}|^2 \) possible interactions. Our hypothesis is that graph topology could be used to detect the areas in the graph where the message passing is failing (i.e., bottlenecks, disconnected components) to reduce the number of long-range interactions that have to be analyzed and avoid the computational inefficiency introduced by transformers.

\textbf{Multi-scale graph architecture.}
Multi-scale models are another popular solution to long-range dependencies detection \cite{dong2023megraph}. They typically implement a hierarchical clustering algorithm to produce several scaled graphs which are then fed into the model. This approach presents two main drawbacks. \emph{First}, it has a high memory footprint and computational complexity. \emph{Second}, they are effective only on specific graph topologies and densities due to the clustering algorithms used. 

\textbf{Topological graph rewiring.} Many rewiring algorithms exploit graph curvature \cite{nguyen2023revisiting} or other topological metrics \cite{barbero2023locality} to identify parts of the graph suffering from over-squashing and over-smoothing and alleviate these issues by adding or removing edges.
\begin{definition}[Over-smoothing]
Over-smoothing refers to the indistinguishable representations of nodes in different classes as the number of layers increases, weakening the expressiveness of deep GNNs and limiting their applicability \cite{chen2020measuring}.
This phenomenon can be formalized as: $\sum_{(u,v) \in \mathcal{V} : y(u) \neq y(v)} |\boldsymbol{X}^{k}_u-\boldsymbol{X}^{k}_v| \rightarrow 0 \text{ as } k \rightarrow \infty$

\label{over-smoothing}
\end{definition}

Where \(k\) refers to the number of layers, \(\boldsymbol{X}\) the feature representations of the nodes, \(y(x)\) to the class of the node \(x\), and \(\mathcal{V}\) the set of nodes.

\begin{definition}[Over-squashing] Over-squashing refers to the exponential growth of a node's receptive field, leading to the collapse of substantial information into a fixed-sized feature vector. As illustrated in \textbf{Fig}~\ref{fig:bottleneck}, the nodes which are close to the bottleneck will have a receptive field which will grow exponentially with the number of layers in the GNN \cite{alon2020bottleneck,di2023over}.
\label{over-squashing}
\end{definition}

\begin{definition}[Homophily] Homophily is the tendency of similar nodes in a network to be connected with each other (i.e., sharing the same label). Its opposite is termed heterophily \cite{ma2021homophily}.
\label{homphily}
\end{definition}

These methods are generally successful at alleviating over-squashing and over-smoothing while preserving the graph topology. However, when dealing with large graphs or specific graph topologies (e.g., chain-like structures), these pre-processing techniques either fail to resolve graph bottlenecks or excessively perturb the graph structure, degrading performance. This degradation is due to the intrinsic limitation of the local neighborhood aggregation paradigm in GNNs. Therefore, to capture LRIs, we must design a more refined aggregation protocol which exploits the graph topology.

\section{Methodology: \textit{DuoGNN}}
In this section we present, DuoGNN, our solution to the following question:
\begin{mdframed}[frametitle={\colorbox{white}{\space Key challenge:\space}},
frametitleaboveskip=-\ht\strutbox,
frametitlealignment=\center
]
How can we design a \emph{scalable} GNN model, which jointly detects short-range and long-range interactions for \emph{any} graph topology?
\end{mdframed}

\begin{figure}[tb]
\centering
\includegraphics[width = 0.82\hsize]{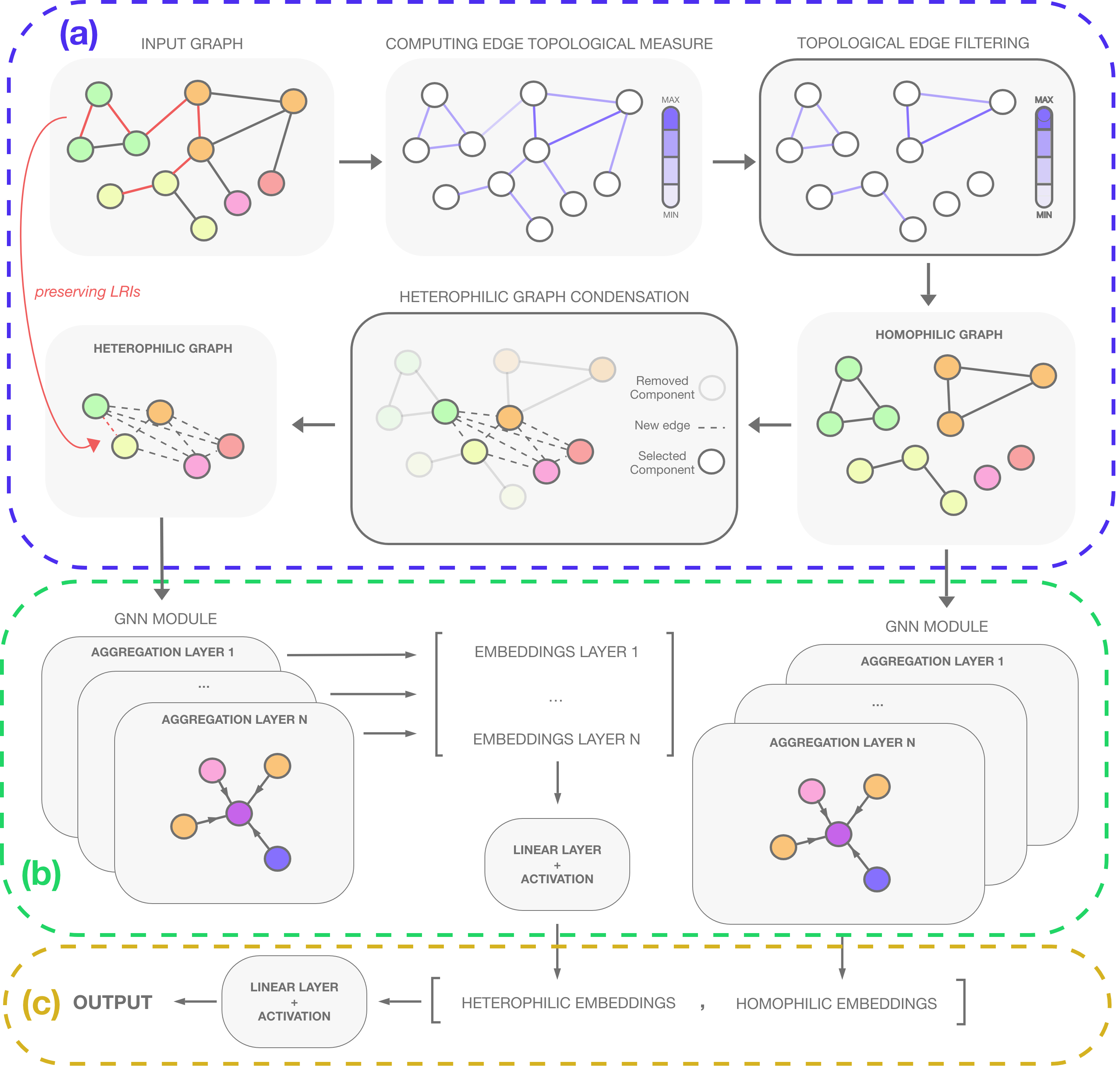}
\caption{The DuoGNN pipeline graph flow includes: (a) An interaction-decoupling stage responsible for extracting the homophilic and heterophilic edges from the input graph. Nodes are colored based on their class, blue-shaded edges indicate the value of the topological measure, and red edges are used to show the LRI preservation throughout the process. (b) A parallel transformation stage that independently processes the two output graphs from the previous stage. Node colors represent node representations. (c) A prediction stage that concatenates the results from the previous stage.}
\label{fig:complete_flow}
\end{figure}

To this end, we propose \textbf{DuoGNN}, a novel GNN model that leverages topological measures to capture short-range and long-range interactions with limited additional time complexity. First, our model breaks the graph into homophilic clusters by removing bottlenecks but maintaining homophilic short-range interactions. However, this process causes the loss of long-range interactions, which we recover using our novel \emph{heterophilic graph condensation}. Next, short-term and long-term interactions are independently learned by a dual aggregation pipeline. An overview of the model pipeline is illustrated in \textbf{Fig}~\ref{fig:complete_flow}. 

\textbf{Interaction-decoupling stage.}
The \emph{interaction-decoupling} stage includes a \emph{topological edge filtering} and a \emph{heterophilic graph condensation} step. This stage is designed to discriminate between homophilic and heterophilic node interactions, producing a strongly homophilic graph and a strongly heterophilic graph. These two graphs are then processed by independent GNN modules in the next stage. This operation is crucial because we want to avoid using a standard GNN aggregation on heterophilic edges to reduce the risk of over-smoothing. The \emph{topological edge filtering} is responsible for preserving homophilic interactions by removing heterophilic edges in the graph and breaking the graph into homophilic connected components. To perform the edge filtering we first compute a topological measure over the edges of the input graph. Next, we proceed by removing the \(\kappa\) least connected edges, since these will likely be close to graph bottlenecks. Here, the definition of connectivity depends on the topological measure applied to the graph, for instance, if we use a centrality measure the least connected edge will have the highest value of the metric. As a result, the output graph of this first step, defined as \(\mathcal{G}_{\textit{ho}}\), will be highly homophilic (\textbf{Figs. 1, 2, 3-supp}). This graph will be resistant against over-smoothing while preserving most short-term node interactions. Next, we proceed with the \emph{heterophilic graph condensation}, which aims at extracting the most relevant heterophilic interactions within the homophilic clusters. To achieve this, we select the most connected node, which will likely have the most reliable representation, for the \(\mu\) most populated cluster of \(\mathcal{G}_{\textit{ho}}\) and build a distinct fully-connected graph with these. The hyper-parameter \(\mu\) is introduced to limit the number of selected clusters in sparse graphs. The resulting graph, defined as \(\mathcal{G}_{\textit{he}}\), will be highly heterophilic --since nodes with different labels will be likely neighbours due to the clustering done during the \emph{topological edge filtering} (\textbf{Figs. 1, 2, 3-supp}). This graph will preserve the majority of the LRIs while resulting considerably smaller than the original graph. This first stage radically differs from previously mentioned graph rewiring algorithms since it does not merely pre-process the graph but classifies how each node interaction should be aggregated based on its homophilic or heterophilic nature. All graph rewiring algorithms proposed in the literature add new edges to relax bottlenecks (increasing information flow) and remove connected edges to reduce over-smoothing (reducing information flow). We instead remove the edges that generate bottlenecks to create distinct connected components and build an additional graph to model them. This approach addresses both over-smoothing and over-squashing with a single operation while introducing significantly less topological perturbation. Additionally, we overcome the limitations of traditional graph rewiring techniques by ensuring that different graph topologies are mapped to distinct sizes of \(\mathcal{G}_{\textit{ho}}\) and \(\mathcal{G}_{\textit{he}}\), and consequently processed differently during the transformation stage.

\textbf{Parallel transformation stage.} In the parallel transformation phase, we independently process \(\mathcal{G}_{\textit{ho}}\) and \(\mathcal{G}_{\textit{he}}\) using distinct GNN modules to enhance model expressiveness. While one module learns to aggregate nodes of similar classes, the other learns to distinguish between dissimilar neighbours. This approach minimizes over-smoothing, as the homophilic module primarily aggregates embeddings of nodes within the same class, and the heterophilic module remains unaffected due to its heterophilic aggregation which we define below. The two graphs are denoted as $\mathcal{G}_{\textit{ho}} = (\mathcal{V}_{\textit{ho}},\mathcal{E}_{\textit{ho}})$ and $\mathcal{G}_{\textit{he}} = (\mathcal{V}_{\textit{he}},\mathcal{E}_{\textit{he}})$, where $\mathcal{V}$ refers to the node set and $\mathcal{E}$ the edge set. The two graphs are fed into two distinct GNN modules. The homophilic module is defined by the well-known formula:
\begin{equation}
\scalebox{0.8}{$
    \boldsymbol{X}_{\textit{ho}}^{(i+1)} = \phi_{\textit{ho}} \left( 
    \hat{\boldsymbol{A}}_{\textit{ho}} \boldsymbol{X}_{\textit{ho}}^{(i)}
    \boldsymbol{W}_{\textit{ho}}^{(i)}\right)  \quad \text{for } 0 \leq i \leq l-1, \quad  \boldsymbol{X}_{\textit{ho}}^{(out)} = \boldsymbol{X}_{\textit{ho}}^{(l)}
    $}
\label{equation_homo}
\end{equation}
where \(\hat{\boldsymbol{A}}_{\textit{ho}} = \boldsymbol{A}_{\textit{ho}} + \boldsymbol{I}\) and  $\boldsymbol{A}_{\textit{ho}} \in \mathbb{R}^{|\mathcal{V}_{\textit{ho}}| \times |\mathcal{V}_{\textit{ho}}|}$ is the adjacency matrix of the homophilic graph, \(\phi_{\textit{ho}}\) is the activation function of the homophilic module, \(\boldsymbol{W}_{\textit{ho}}^{(i)}\) a learnable matrix of weights of the layer \(i\) and \(\boldsymbol{X}_{\textit{ho}}^{(i)}\) the node embeddings of the layer \(i\). For the heterophilic aggregation, we alter this formulation to allow the model to learn how to distinguish distinct node classes and prevent the over-smoothing of the feature representations. Firstly, we do not aggregate the node features in the first layer so that the model can process the original features (see equation \ref{equation_hete}). Additionally, we build a row vector of all layers' node embedding outputs (see equation \ref{equation_out}) and feed it into a linear layer \cite{xu2018representation}. This ensures that the model learns to distinguish between different node classes at many levels of smoothness.
\begin{equation}
\scalebox{0.80}{$
    \boldsymbol{X}_{\textit{he}}^{(1)} = \phi_{\textit{he}} \left( 
    \boldsymbol{X}_{\textit{he}}^{(0)}
    \boldsymbol{W}_{\textit{he}}^{(0)}\right) \quad \text{,} \quad \boldsymbol{X}_{\textit{he}}^{(i+1)} = \phi_{\textit{he}} \left( 
    \boldsymbol{A}_{\textit{he}}
    \boldsymbol{X}_{\textit{he}}^{(i)}
    \boldsymbol{W}_{\textit{he}}^{(i)}+ 
    \boldsymbol{X}_{\textit{he}}^{(i)}
    \hat{\boldsymbol{W}}_{\textit{he}}^{(i)}\right)
    $}
\label{equation_hete}
\end{equation}

where \(\hat{\boldsymbol{W}}_{\textit{he}}^{(i)}\) is a learnable weight matrix used to process the self-connections and the remaining symbols are the heterophily equivalent to what has been illustrated in equation \ref{equation_homo}. While other previous works have proposed a parallel homophilic and heterophilic aggregation \cite{du2022gbk}, our novel topological interaction-decoupling mechanism eliminates the need for additional neural components to make this distinction. This results in a more efficient and simpler model architecture when compared to these approaches.

\textbf{Prediction stage.} Finally, we concatenate and process both modules' outputs through a final linear layer to perform the prediction (see equation \ref{equation_out}).
\begin{equation}
\scalebox{0.7}{$
\boldsymbol{X}_{\textit{he}}^{(out)} = \phi_{\textit{he}}\left(
\boldsymbol{W}_{\textit{he}}^{(out)}
\begin{bmatrix}
\boldsymbol{X}_{\textit{he}}^{(1)} \\
\boldsymbol{X}_{\textit{he}}^{(2)} \\
\vdots \\
\boldsymbol{X}_{\textit{he}}^{(l)}
\end{bmatrix}\right) \text{, }
\boldsymbol{X}^{(out)} = \phi_{\textit{out}}\left(
\boldsymbol{W}^{(out)}
\begin{bmatrix}
\boldsymbol{X}_{\textit{ho}}^{(out)} \\
\boldsymbol{X}_{\textit{he}}^{(out)}
\end{bmatrix}\right)
$}
\label{equation_out}
\end{equation}

\section{Results and Discussion}

\textbf{MedMNIST Organ C \& Organ S dataset and hyperparameter setting.}  The MedMNIST Organ C and Organ S datasets \cite{medmnistv2} include abdominal CT scan images (28 $\times$ 28 pixels) of liver tumours, in coronal and sagittal views respectively. The task is node-level multi-class classification among 7 types of liver tumours. To process the images with a GNN model we convert them to a graph by representing each image as a node with vector embedding of length 784 (vectorisation of the 28 by 28 pixels intensity). The edges of the graph are derived from the cosine similarity of the node embeddings. They are sparsified to the top $\sim$1.2 million edges and then the graph is converted to an unweighted graph to reduce its complexity, similar to what was done in \cite{adnel2023affordable}. The hyper-parameters of DuoGNN were set to \(\kappa = 50000\) and \(\mu = 500\) following a grid-search conducted on the validation set. These values are considerably higher than those used for the Cora dataset due to the elevated number of edges and connected components. A summary of the datasets along the hyper-parameters used is shown in \textbf{Table 1-supp}.

\textbf{Cora dataset and hyperparameter setting.} We further evaluate our framework on the Cora \cite{mccallum2000automating} dataset. The dataset includes one graph, where the nodes denote scientific publications and the edges citations. The node embeddings are real-valued vectors of length 1433 (size of the dictionary) describing the paper, each value indicates the presence of a word in the dictionary. The task is node-level multi-class classification among 7 fields of research. The hyper-parameters of DuoGNN were set to \(\kappa = 60\) and \(\mu = 20\) following a grid-search conducted on the validation set.

\textbf{Dataset benchmark results and discussion.} We benchmarked DuoGNN against common GNN baselines such as GCN \cite{kipf2016semi}, GIN \cite{xu2018powerful}, and GAT \cite{GAT}. We implemented 4 variants of DuoGNN implementing distinct topological measures (all using GCN aggregation layers) and 3 ablations integrating the \emph{topological edge filtering} on the baselines. The variants consistently outperformed the baselines by a considerable margin across almost all datasets (\textbf{Table \ref{tab:benchmark_comparison}}). This demonstrates the model's enhanced capability to distinguish between distinct node labels in a node classification setting and its robustness, as evidenced by its consistent performance across various topological metrics. The highest performing DuoGNN variant was the curvature-enhanced version, which made use of the discrete Olivier's Ricci curvature \cite{lin2011ricci}. This could be explained by its proven theoretical ability to detect areas in the graph affected by over-smoothing or over-squashing \cite{topping2021understanding}. The proposed ablations showed marginal improvements when compared to the baselines, proving that simple pre-processing techniques are not expressive enough to capture LRIs. DuoGNN showed great topological generalizability thanks to the flexibility of the parameters \(\kappa\) and \(\mu\) which can fit any graph density and structure. 

\begin{table*}[htbp]
  \centering
  \caption{Prediction result comparison of various methods.}
  \scalebox{0.8}{
    \begin{tabular}{l|cc|cc|cc}
    \multirow{2}[1]{*}{Methods} & \multicolumn{2}{c|}{CORA} & \multicolumn{2}{c|}{MedMNIST Organ-S} & \multicolumn{2}{c}{MedMNIST Organ-C} \\
          & ACC   & Specificity    & ACC   & Specificity & ACC   & Specificity \\
    \midrule
    GCN   & 84.37±0.76      & 97.39±0.12     & 60.12±0.08 & 96.01±0.01 & 77.68±0.35 & 97.77±0.04 \\
    GCN \textit{+ filtering (curvature)} & 85.07±0.70  & 97.51±0.11  & 58.98±0.41 & 95.90±0.04 & 77.15±0.29 & 97.71±0.03 \\
    GIN   & 83.47±0.46      &   97.24±0.06   & 60.36±0.00 & 96.07±0.00 & 76.33±0.00 & 97.63±0.00 \\
    GIN \textit{+ filtering (curvature)} & 83.90±0.65  & 97.32±0.10  & 61.51±0.47 & 96.15±0.05 & 77.26±1.47 & 97.72±0.48 \\
    GAT & \cellcolor[HTML]{9CEB85} 85.87±0.32  & \cellcolor[HTML]{9CEB85} 97.65±0.05  & OOM & OOM & OOM & OOM \\
    GAT \textit{+ filtering (curvature)} & 85.35±0.83  &  97.56±0.14  & OOM & OOM & OOM & OOM \\
    DuoGNN \textit{random} & \cellcolor[HTML]{E7FFE0} 85.70±0.61  & \cellcolor[HTML]{E7FFE0} 97.62±0.10     & 62.63±0.31 & 96.26±0.03 & 79.39±0.90 & 97.93±0.09 \\
    DuoGNN \textit{eigen} & 85.20±1.01    & 97.53±0.16     & \cellcolor[HTML]{9CEB85} 62.66±0.24 & \cellcolor[HTML]{9CEB85} 96.26±0.02 & \cellcolor[HTML]{9CEB85} 80.22±1.14 & \cellcolor[HTML]{9CEB85} 98.02±0.11 \\
    DuoGNN \textit{degree} &  85.32±0.12  & 97.55±0.02       &  \cellcolor[HTML]{E7FFE0} 62.64±0.22 & \cellcolor[HTML]{E7FFE0} 96.26±0.02 & \cellcolor[HTML]{E7FFE0} 79.73±1.05 & \cellcolor[HTML]{E7FFE0} 97.97±0.10 \\
    DuoGNN \textit{curvature} & \cellcolor[HTML]{56EB2A} \textbf{85.91±0.49} & \cellcolor[HTML]{56EB2A} \textbf{97.65±0.04}     & \cellcolor[HTML]{56EB2A} \textbf{63.06±0.42} & \cellcolor[HTML]{56EB2A} \textbf{96.30±0.04} & \cellcolor[HTML]{56EB2A} \textbf{80.27±0.53} & \cellcolor[HTML]{56EB2A} \textbf{98.02±0.05} \\
    \end{tabular}%
    }
  \label{tab:benchmark_comparison}%

\begin{minipage}{2\columnwidth}
\vspace{0.1cm}
\small Notes: Results better than their counterparts have a more intense shade of green.
\end{minipage}
\end{table*}%

Moreover, DuoGNN demonstrates to be scalable thanks to its \emph{heterophilic graph condensation} which reduces dramatically the number of LRIs which have to be analyzed. In general, since the number of clusters is considerably smaller than the number of nodes in the graph, we can conclude that DuoGNN has the same time complexity as the aggregation layer used in the \emph{parallel transformation} stage. DuoGNN shows a higher GPU memory footprint when processing small graphs thanks to its dual aggregation paradigm, which inherently requires more memory to process a single batch. However, it becomes more scalable than attention-based models such as GAT for large graphs thanks to its \textit{heterophilic graph condensation}  (\textbf{Table}~\ref{tab:main_resource}).

\begin{table*}[ht!]
  \centering
  \caption{Computational resources comparison of baselines and proposed model.}
  \scalebox{0.8}{
    \begin{tabular}{p{13.6em}|>{\centering\arraybackslash}p{5.7em}>{\centering\arraybackslash}p{5em}|>{\centering\arraybackslash}p{5.7em}>{\centering\arraybackslash}p{5em}|>{\centering\arraybackslash}p{5.7em}>{\centering\arraybackslash}p{5em}}
    \multirow{2}[0]{*}{Methods} & \multicolumn{2}{c|}{CORA} & \multicolumn{2}{c|}{MedMNIST Organ-S} & \multicolumn{2}{c}{MedMNIST Organ-C} \\
    \multicolumn{1}{c|}{} & GPU Memory & Epoch Time & GPU Memory & Epoch Time & GPU Memory & Epoch Time \\
    \midrule
    GCN \textit{(and variant)} & \textbf{148.6} & \textbf{0.16±0.0} & 1175.4 & \textbf{1.7±0.0} & 1102.4 & \textbf{1.9±0.1} \\
    GIN \textit{(and variant)} & 340.7 & 0.24±0.0 & \textbf{960.7} & 2.5±0.1 & \textbf{896.6} & 2.4±0.0 \\
    GAT \textit{(and variant)} & 414.2 & 0.21±0.0 & OOM & OOM & OOM & OOM \\
    DuoGNN \textit{(all variants)} & 514.6 & 0.29±0.0 & 1451.2 & 3.1±0.0 & 1358.8 & 3.2±0.0 \\ 
    \end{tabular}%
    }
  \label{tab:main_resource}%
\begin{minipage}{2\columnwidth}
\vspace{0.1cm}
\vspace{0.1cm}
\small\raggedright Notes: Epoch time is defined as the time for one complete pass of the dataset. The GPU \\ memory is in MB, and the epoch time is in milliseconds (overall lowest is \textbf{bolded}).
\end{minipage}
\end{table*}%

\vspace{-5mm}

\section{Conclusion}
In this work, we introduced DuoGNN, a novel topology-aware GNN architecture for jojntly detecting short-range and long-range interactions. Our core contributions include: (1) a topological edge-filtering algorithm, and (2) a heterophilic graph condensation technique that ensures scalability and generalizability. Additionally, we designed a parallel homophilic and heterophilic aggregation mechanism to independently process both types of interactions. Our benchmarking results demonstrate that DuoGNN outperforms the baselines by a considerable margin. As a future direction, we aim to better understand the role of different topological metrics in alleviating over-smoothing and over-squashing issues and to develop a hyperparameter-free version of the model.



\newpage
\section{Supplementary material}
\begin{figure}[htbp]
    \centering
    \includegraphics[width=0.85\textwidth]{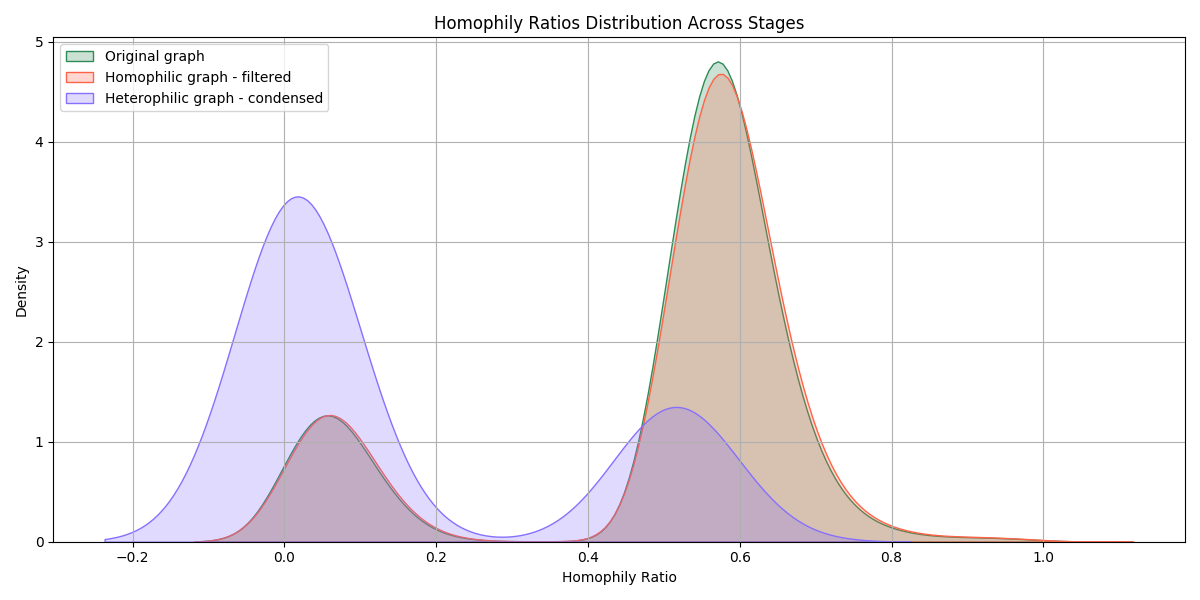}
    \caption{Homophily ratio distribution shift during the interaction-decoupling stage (Cora Dataset)}
    \label{density1}
\end{figure}

\begin{figure}[htbp]
    \centering
    \includegraphics[width=0.85\textwidth]{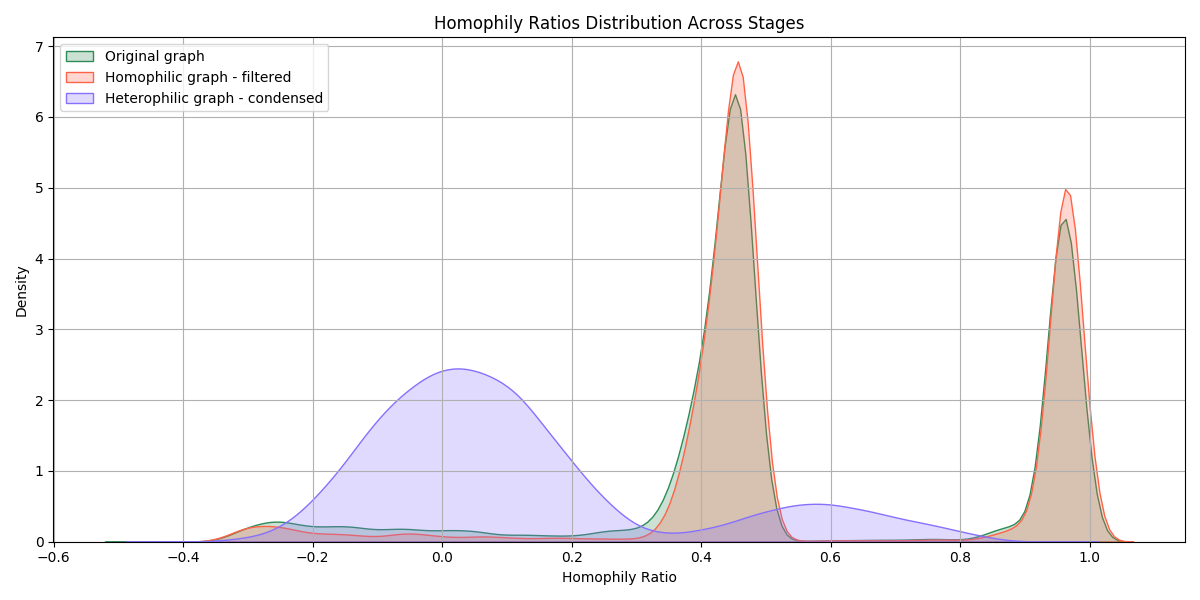}
    \caption{Homophily ratio distribution shift during the interaction-decoupling stage (Organ-S Dataset)}
    \label{density2}
\end{figure}

\begin{figure}[htbp]
    \centering
    \includegraphics[width=0.85\textwidth]{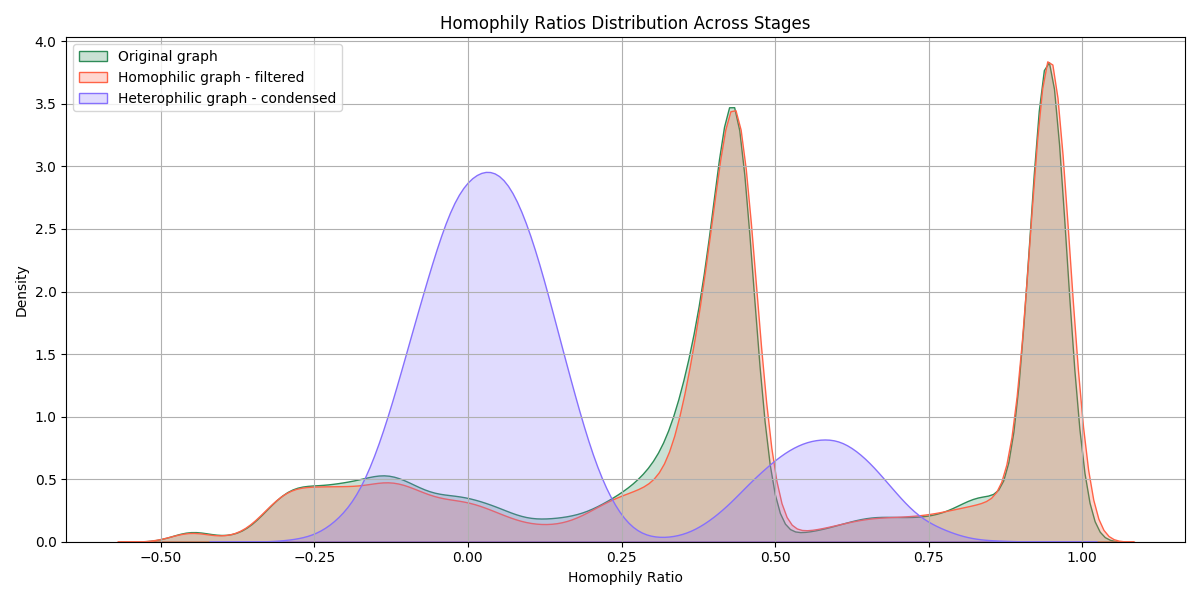}
    \caption{Homophily ratio distribution shift during the interaction-decoupling stage (Organ-C Dataset)}
    \label{density3}
\end{figure}

We provide three supplementary materials to facilitate reproducibility and offer additional insights into DuoGNN:

\begin{itemize}
  \item Dataset and hyperparameters details (see Table \ref{tab:dataset_details}). Homophily ratio distribution plots during the interaction-decoupling stage (see Figures \ref{density1}-\ref{density3}).
  \item A 7-mn YouTube video explaining how DuoGNN works on the BASIRA YouTube channel at \href{https://youtu.be/6WtlQQDanaQ}{https://youtu.be/6WtlQQDanaQ}.
  \item DuoGNN code in Python on GitHub at  
  \href{https://github.com/basiralab/DuoGNN}{https://github.com/basiralab/DuoGNN}.
\end{itemize}

\begin{table}[htbp]
  \centering
  \caption{Dataset details and hyperparameters.}
  \scalebox{0.83}{
    \begin{tabular}{l|c|c|c}
          & \textbf{Organ-S} & \textbf{Organ-C} & Cora \\
    \midrule
    \# of Nodes & 25221 & 23660 & 2708 \\
    \# of Edges & 1276046 & 1241622 & 5278 \\
    \# of Features & 784 & 784 & 1433 \\
    \# of Labels & 11 & 11 & 7 \\
    Task Type & Multi-class & Multi-class & Multi-class \\
    Training Type & Inductive & Inductive & Inductive \\
    \midrule
    Training Nodes & 13940 & 13000 & 1208 \\
    Validation Nodes & 2452 & 2392 & 500 \\
    Test Nodes & 8829 & 8268 & 1000 \\
    \midrule
    \# of layers & 3 & 3 & 3 \\
    Hidden channels & 1024 & 1024 & 2048 \\
    GNN module (DuoGNN) & GCN & GCN & GCN \\
    Dropouts & 0.5 & 0.5 & 0.5 \\
    \(\kappa\) (\# removed edges - filtering) & 50000 & 50000 & 60 \\
    \(\mu\) (\# max communities - filtering) & 500 & 500 & 20 \\
    Learning Rate & 0.0005 & 0.0005 & 0.0005 \\
    \end{tabular}%
    }
  \label{tab:dataset_details}%
\end{table}%

\newpage

\begin{thebibliography}{10}

\bibitem{scarselli2008graph}
Scarselli, F., Gori, M., Tsoi, A.C., Hagenbuchner, M., Monfardini, G.:
\newblock The graph neural network model.
\newblock IEEE transactions on neural networks \textbf{20} (2008)  61--80

\bibitem{NPMS}
Gilmer, J., Schoenholz, S.S., Riley, P.F., Vinyals, O., Dahl, G.E.:
\newblock Message passing neural networks (2020)

\bibitem{PPI}
Zitnik, M., Leskovec, J.:
\newblock Predicting multicellular function through multi-layer tissue networks.
\newblock CoRR \textbf{abs/1707.04638} (2017)

\bibitem{xu2018powerful}
Xu, K., Hu, W., Leskovec, J., Jegelka, S.:
\newblock How powerful are graph neural networks?
\newblock arXiv preprint arXiv:1810.00826 (2018)

\bibitem{surveyGNN}
Wu, Z., Pan, S., Chen, F., Long, G., Zhang, C., Yu, P.S.:
\newblock A comprehensive survey on graph neural networks.
\newblock CoRR \textbf{abs/1901.00596} (2019)

\bibitem{cui2021metro}
Cui, Y., Zheng, K., Cui, D., Xie, J., Deng, L., Huang, F., Zhou, X.:
\newblock Metro: a generic graph neural network framework for multivariate time series forecasting.
\newblock Proceedings of the VLDB Endowment \textbf{15} (2021)  224--236

\bibitem{GCNmedical}
Ahmedt-Aristizabal, D., Armin, M.A., Denman, S., Fookes, C., Petersson, L.:
\newblock Graph-based deep learning for medical diagnosis and analysis: Past, present and future.
\newblock Sensors \textbf{21} (2021)  4758

\bibitem{basira-neuroscience}
Bessadok, A., Mahjoub, M.A., Rekik, I.:
\newblock Graph neural networks in network neuroscience (2015)

\bibitem{medmnistv2}
Yang, J., Shi, R., Wei, D., Liu, Z., Zhao, L., Ke, B., Pfister, H., Ni, B.:
\newblock Medmnist v2-a large-scale lightweight benchmark for 2d and 3d biomedical image classification.
\newblock Scientific Data \textbf{10} (2023) ~41

\bibitem{qureshi2023limits}
Qureshi, S.,  et~al.:
\newblock Limits of depth: Over-smoothing and over-squashing in gnns.
\newblock Big Data Mining and Analytics \textbf{7} (2023)  205--216

\bibitem{rusch2023survey}
Rusch, T.K., Bronstein, M.M., Mishra, S.:
\newblock A survey on oversmoothing in graph neural networks.
\newblock arXiv preprint arXiv:2303.10993 (2023)

\bibitem{fei2023gnn}
Fei, Z., Guo, J., Gong, H., Ye, L., Attahi, E., Huang, B.:
\newblock A gnn architecture with local and global-attention feature for image classification.
\newblock IEEE Access (2023)

\bibitem{wu2021representing}
Wu, Z., Jain, P., Wright, M., Mirhoseini, A., Gonzalez, J.E., Stoica, I.:
\newblock Representing long-range context for graph neural networks with global attention.
\newblock Advances in Neural Information Processing Systems \textbf{34} (2021)  13266--13279

\bibitem{nguyen2023revisiting}
Nguyen, K., Hieu, N.M., Nguyen, V.D., Ho, N., Osher, S., Nguyen, T.M.:
\newblock Revisiting over-smoothing and over-squashing using ollivier-ricci curvature.
\newblock In: International Conference on Machine Learning, PMLR (2023)  25956--25979

\bibitem{barbero2023locality}
Barbero, F., Velingker, A., Saberi, A., Bronstein, M., Di~Giovanni, F.:
\newblock Locality-aware graph-rewiring in gnns.
\newblock arXiv preprint arXiv:2310.01668 (2023)

\bibitem{dong2023megraph}
Dong, H., Xu, J., Yang, Y., Zhao, R., Wu, S., Yuan, C., Li, X., Maddison, C.J., Han, L.:
\newblock Megraph: Capturing long-range interactions by alternating local and hierarchical aggregation on multi-scaled graph hierarchy.
\newblock Advances in Neural Information Processing Systems \textbf{36} (2023)  63609--63641

\bibitem{chen2023multi}
Chen, L., Chen, D., Shang, Z., Wu, B., Zheng, C., Wen, B., Zhang, W.:
\newblock Multi-scale adaptive graph neural network for multivariate time series forecasting.
\newblock IEEE Transactions on Knowledge and Data Engineering (2023)

\bibitem{chen2020measuring}
Chen, D., Lin, Y., Li, W., Li, P., Zhou, J., Sun, X.:
\newblock Measuring and relieving the over-smoothing problem for graph neural networks from the topological view.
\newblock In: Proceedings of the AAAI conference on artificial intelligence. Volume~34. (2020)  3438--3445

\bibitem{alon2020bottleneck}
Alon, U., Yahav, E.:
\newblock On the bottleneck of graph neural networks and its practical implications.
\newblock arXiv preprint arXiv:2006.05205 (2020)

\bibitem{di2023over}
Di~Giovanni, F., Giusti, L., Barbero, F., Luise, G., Lio, P., Bronstein, M.M.:
\newblock On over-squashing in message passing neural networks: The impact of width, depth, and topology.
\newblock In: International Conference on Machine Learning, PMLR (2023)  7865--7885

\bibitem{ma2021homophily}
Ma, Y., Liu, X., Shah, N., Tang, J.:
\newblock Is homophily a necessity for graph neural networks?
\newblock arXiv preprint arXiv:2106.06134 (2021)

\bibitem{xu2018representation}
Xu, K., Li, C., Tian, Y., Sonobe, T., Kawarabayashi, K.i., Jegelka, S.:
\newblock Representation learning on graphs with jumping knowledge networks.
\newblock In: International conference on machine learning, PMLR (2018)  5453--5462

\bibitem{du2022gbk}
Du, L., Shi, X., Fu, Q., Ma, X., Liu, H., Han, S., Zhang, D.:
\newblock Gbk-gnn: Gated bi-kernel graph neural networks for modeling both homophily and heterophily.
\newblock In: Proceedings of the ACM Web Conference 2022. (2022)  1550--1558

\bibitem{adnel2023affordable}
Adnel, C., Rekik, I.:
\newblock Affordable graph neural network framework using topological graph contraction.
\newblock In: Workshop on Medical Image Learning with Limited and Noisy Data, Springer (2023)  35--46

\bibitem{mccallum2000automating}
McCallum, A.K., Nigam, K., Rennie, J., Seymore, K.:
\newblock Automating the construction of internet portals with machine learning.
\newblock Information Retrieval \textbf{3} (2000)  127--163

\bibitem{kipf2016semi}
Kipf, T.N., Welling, M.:
\newblock Semi-supervised classification with graph convolutional networks.
\newblock arXiv preprint arXiv:1609.02907 (2016)

\bibitem{GAT}
Veličković, P., Cucurull, G., Casanova, A., Romero, A., Liò, P., Bengio, Y.:
\newblock Graph attention networks (2018)

\bibitem{lin2011ricci}
Lin, Y., Lu, L., Yau, S.T.:
\newblock Ricci curvature of graphs.
\newblock Tohoku Mathematical Journal, Second Series \textbf{63} (2011)  605--627

\bibitem{topping2021understanding}
Topping, J., Di~Giovanni, F., Chamberlain, B.P., Dong, X., Bronstein, M.M.:
\newblock Understanding over-squashing and bottlenecks on graphs via curvature.
\newblock arXiv preprint arXiv:2111.14522 (2021)

\end{thebibliography}
\end{document}